\documentclass{article}
\usepackage[ruled]{algorithm2e}
\usepackage{amsmath}
\usepackage{amsfonts}
\usepackage{stmaryrd}
\usepackage{textcomp}
\usepackage{caption}
\usepackage{url}
\usepackage{graphicx}
\usepackage{multirow,hhline}
\usepackage{subfig} 
\usepackage{fullpage}

\usepackage{tabularx} 
\newcolumntype{Y}{>{\raggedright\arraybackslash}X}

\title{Understanding the Mechanisms of Deep Transfer Learning for Medical Images}
\author{Hariharan Ravishankar, Prasad Sudhakar, \\ Rahul Venkataramani, Sheshadri Thiruvenkadam, Pavan Annangi, \\
Narayanan Babu and Vivek Vaidya\\\\
\emph{GE Global Research, Bangalore, India}}
\date{}
\begin{document}
	\maketitle
\begin{abstract}
The ability to automatically learn task specific feature representations has led to a huge success of deep learning methods. When large training data is scarce, such as in medical imaging problems, transfer learning has been very effective. In this paper, we systematically investigate the process of transferring a Convolutional Neural Network, trained on ImageNet images to perform image classification, to kidney detection problem in ultrasound images. We study how the detection performance depends on the extent of transfer. We show that a transferred and tuned CNN can outperform a state-of-the-art feature engineered pipeline and a hybridization of these two techniques achieves 20\% higher performance. We also investigate how the evolution of intermediate response images from our network. Finally, we compare these responses to state-of-the-art image processing filters in order to gain greater insight into how transfer learning is able to effectively manage widely varying imaging regimes.
\end{abstract}	


\section{Introduction}
Automated organ localization and segmentation from ultrasound images is a challenging problem because of specular noise, low soft tissue contrast and wide variability of data from patient to patient. In such difficult problem settings, data driven machine learning methods, and especially deep learning methods in recent times, have found quite a bit of success. Usually, a large amount of labeled data is needed to train machine learning models and a careful
feature engineering is required for each problem. The question of how much data is needed for satisfactory performance of these methods is still unanswered, with some recent works in this direction~\cite{Cho:2016}. However,~\emph{transfer learning} has been successfully employed in data scarce situations, with model knowledge being effectively transferred across (possibly unrelated) tasks/domains. It is fascinating that a model, learnt for an unrelated problem setting can actually solve a problem at hand with minimal retraining. 
	In this paper, we have attempted to demonstrate and understand the effectiveness and mechanism of transfer learning a CNN,  originally learnt on camera images for image recognition, to solve the problem of automated kidney localization from ultrasound B-mode images.
	
	 Kidney detection is challenging due to wide variability in its shape, size and orientation. Depending upon the acquisition scan plane, inconsistency in appearance of internal regions (renal sinus) and presence of adjacent structures like diaphragm, liver boundaries, etc. pose additional challenges. This is also a clinically relevant problem as kidney morphology measurements are essential in assessing renal abnormalities~\cite{Kop:2010}, planning and monitoring radiation therapy, and renal transplant. 

There have been semi-automated and automated kidney detection approaches reported in literature. 
In~\cite{Xie:2005}, a texture model is built by an expectation maximization algorithm using features inferred from a bank of Gabor filters, followed by iterative segmentation to combine texture measures into parametric shape model. In~\cite{Fernandez:2005}, Markov random fields and active contour methods have been used to detect kidney boundaries in 3D ultrasound images. Recently, machine learning approaches~\cite{Ardon:2015,Ravishankar:2016} based on kidney texture analysis have proven successful for segmentation of kidney regions from 2-D and 3-D ultrasound images.


\section{State of the art}

 CNNs \cite{Krizhevsky:2012} provide effective models for vision learning tasks by incorporating spatial context and weight sharing between pixels. 
A typical deep CNN for a learning task has, as input, $N$ channel image patches $P_k$ of size $n_1\times n_2$, where $P_k:\{1,2,\cdots,n_1\} \times \{1,2,\cdots,n_2\}  \rightarrow D\subset \mathbb{R}$, $k=1,2,\cdots,N$. The output is $M$ feature maps, $G_j\in\mathbb{R}^{m_1\times m_2}, j=1,2,\cdots,M$, defined as convolutions using $MN$ filters $v^j_k, (j=1,2,\cdots,M)$, $(k=1,2,\cdots,N) $ of size $S=s_1 \times s_2$, and $M$ scalars $b^j, j=1,2,\cdots, M$. We then have:
\begin{equation}
\label{eqn:CNN} 
G_j=\mathcal{S}_\downarrow\left(\sigma\left(\sum_k P_k * v^j_k +b^j\right)\right), j=1,2,\cdots,M.
\end{equation}
Here, $*$ denotes convolution, $\sigma$ is a non-linear function (sigmoid or a linear cutoff (ReLU)). $\mathcal{S}_\downarrow$ is a down sampling operator. 
The number of feature maps, filter size, and size of the feature maps are hyperparameters in the above expression, with a total of $M(NS+1)$ parameters that one has to optimize for a learning task. A deep CNN architecture is multi-layered, with the above expression being hierarchically stitched together, given the number of input/output maps, sizes of filters and maps for each layer, resulting in a huge number of parameters to be optimized. When data is scarce, the learning problem is under-determined and therefore transferring CNN parameters from a pre-learned model helps.

For medical image problems, transfer learning is additionally attractive due to the heterogeneity of data types (modalities, anatomies, etc.) and clinical challenges. In~\cite{Carneiro:2015}, the authors perform breast image classification using a CNN model trained on ImageNet. Shie et al.~\cite{Shie:2015} employ the CaffeNet, trained on ImageNet, to extract features and classify Otitis Media images. In~\cite{Chen:2015}, a pre-trained CNN is used to extract features on ultrasound images to localize a certain standard plane that is important for diagnosis. 

Studies on transferability of features across CNNs include~\cite{Yosinski:2014} and more specifically~\cite{Roth:2016},\cite{Tajbakhsh:2016} for medical images. While our work demonstrates yet another success of transfer learning for medical imaging and the tuning aspects of transfer learning, we 

\begin{enumerate}
	
	\item{Reason out the effectiveness of transfer learning by methodically comparing the response maps from various layers of transfer learnt network with traditional image processing filters.}
		
	\item {Investigate the effect of level of tuning on performance. We demonstrate that full network adaptation leads to learning problem specific features and also establishes the superiority over off-the-shelf image processing filters.}

	\item{Re-establish the relevance and complementary advantages of state-of-the-art, hand-crafted features and merits of hybridisation approaches with CNNs, to help us achieve next level performance improvement~\cite{zheng20153d}.}
	
\end{enumerate}


\section{Methods}
From a set of training images, we build classifiers to differentiate between kidney and non-kidney regions. On a test image, the maximum likelihood detection problem of finding the best kidney region of interest (ROI) $S^*$  from a set of candidate ROIs $\{S\}$ is split into two steps, similar to~\cite{Carneiro:2008,Ravishankar:2016}. The entire set $\{S\}$ is passed through our classifier models and the candidates with positive class labels ($Y$) are retained (Eq.~\eqref{eqn:mlmodel}). 
The ROI with highest likelihood ($L$) from the set $\{S^+\}$ is selected as the detected kidney region (Eq.~\eqref{eqn:likelihood})

\begin{equation}
{\{Y, L\}  = ML Classifier(S) \  \mathrm{and} \ \{S^+ \in S \mid Y = 1\}},
\label{eqn:mlmodel}
\end{equation}
\begin{equation}
\label{eqn:likelihood}
S^* = \arg \max(L^+), \ \mathrm{where} \ \{L^+ = L(S^+)\}.
\end{equation}

We propose to employ CNNs as feature extractors similar to~\cite{Girshick:2013} to facilitate comparisons with traditional texture features. We also propose to use a well-known machine learning classifier, to evaluate performance of different feature sets, thereby eliminating the effects of having soft-max layer for CNNs and a different classifier on traditional features as our likelihood functions. 
\subsection{Dataset and Training}
\label{Training}

We considered a total of 90 long axis kidney images acquired on GE Healthcare LOGIQ E9 scanner, split into two equal and distinct sets, for training  and validation. 
The images contained kidney of different sizes with lengths varying between 7.5cm and 14cm and widths varying between 3.5cm and 7cm, demonstrating wide variability in the dataset.  The orientation of the kidneys varied between -25$^\circ$ and +15$^\circ$. The images were acquired at varying depths of ultrasound acquisition ranging between 9cm and 16cm. Accurate rectangular ground truth kidney ROIs were manually marked by a clinical expert. 

\begin{figure}[!ht]
					\subfloat[ \label{subfig:sample}]{%
					
					\includegraphics[width = 0.32\textwidth, height = 0.28\textwidth]{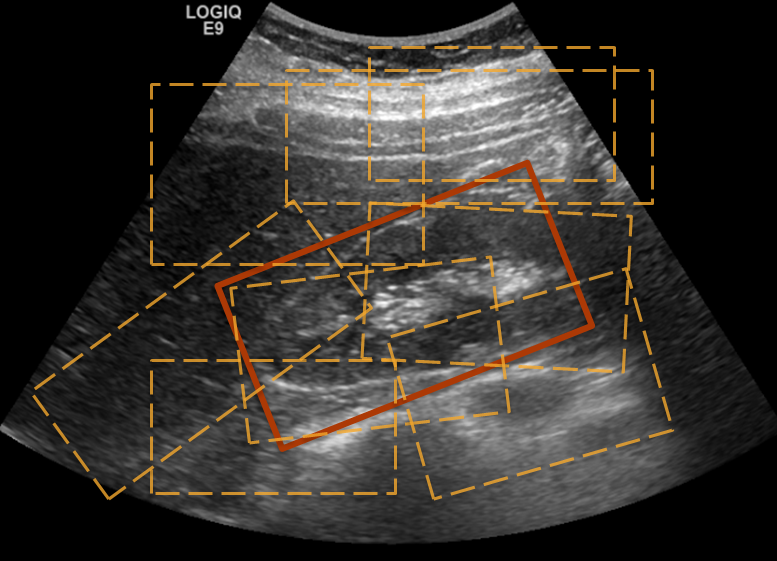}
				}
				\subfloat[\label{subfig:positive}]{%
					\includegraphics[width = 0.32\textwidth, height = 0.28\textwidth]{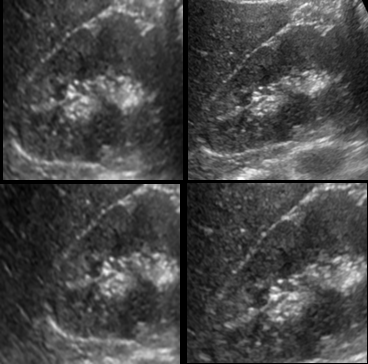}
					
				}	
				\subfloat[\label{subfig:negative}]{%
					\includegraphics[width = 0.32\textwidth, height = 0.28\textwidth]{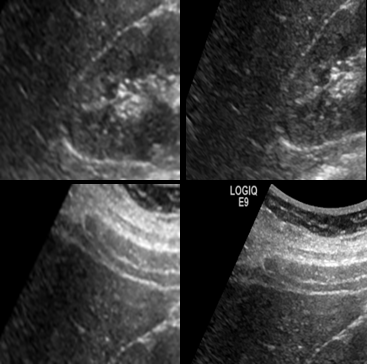}
					
				}\\          
				
				\caption{(a) Sample Generation (b) Positive patches (c) Negative patches }
								\label{fig:patches}
			\end{figure}

To build our binary classification models from training images, we swept the field of view (FOV) to generate many overlapping patches of varying sizes (see Fig. \ref{fig:patches}) that satisfied clinical guidelines on average adult kidney dimensions and aspect ratio~\cite{Emamian:1993}.
We downsampled these ROIs to a common size and were further binned into two classes based on their overlap with ground truth  annotations.  
We used Dice similarity coefficient (DSC) as the metric and a threshold of 0.8 (based on visual and clinical feedback) was used to generate positive and negative class samples.  
This was followed by feature extraction and model building.

\subsection{Transfer Learned Features}
\label{tran}
\begin{figure*}[t] 
				\centering 
				\includegraphics[clip=true,width=1\textwidth]{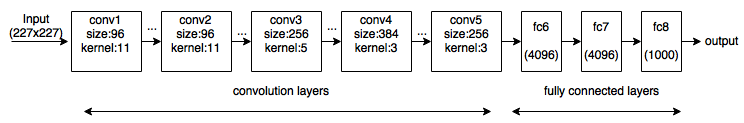} 
				\caption{Simple schematic of CaffeNet architecture [40]}    
				\captionsetup{justification=centering}
				\label{fig:caffenet}	  
			\end{figure*}

Our study on transfer learning was based on adapting the popular CaffeNet~\cite{Jia:2014} architecture built on ImageNet database to ultrasound kidney detection, whose simplified schematic is in Fig.~\ref{fig:caffenet}. We extracted features after the `fc7' layer from all the updated nets, resulting in 4096 features. The features extracted were:
\begin{enumerate}
\item {\textbf{Full Network adaptation (CaffeNet\_FA)} - Initialized with weights from CaffeNet parameters, the entire network weights were updated by training on kidney image samples from Section \ref{Training}. The experiment settings were: stochastic gradient descent update with a batch size of 100, momentum of 0.5 and weight decay of $5\times 10^{-4}$. 

}
\item{\textbf{Partial Network adaptation (CaffeNet\_PA)} - To understand the performance difference based on level of tuning, we froze the weights of `conv1' and `conv2' layers, while updating the weights of other layers. The reasoning behind freezing the first two layers was to evaluate how sharable were the low-level features and also to help us in interpret-ability (Sec. \ref{discuss}).  The experiment settings were same as those for full network adaptation.}
\item{\textbf{Zero Network adaptation (CaffeNet\_NA)} - Finally, we also extracted features from the original CaffeNet model without modifying the weights.}
\end{enumerate}

\subsection{Traditional Texture Features}
\label{trad}

Some of the well-studied texture features used for ultrasound images  include (i) Haar features~\cite{Carneiro:2008} for fetal anatomy studies, (ii) Gray Level Co-Occurrence Matrix (GLCM)~\cite{Sohail:2010},  (iii) Histogram of oriented gradient (HoG) (for automatic view classification of echocardiogram images~\cite{Keramidas:2007}).

Haar features have been reported to have the best performance for kidney detection n~\cite{Ravishankar:2016}.
For our study, we extracted Haar features similar to~\cite{Carneiro:2008}, yielding a total of $\sim$2000 features.

\subsection{Gradient Boosting Machine (GBM)}
Ensemble classifiers have been shown to be successful in ultrasound organ detection problems. In~\cite{Carneiro:2008}, authors have used probabilistic boosting tree classifier for fetal anatomy detection. In~\cite{Ravishankar:2016}, it has been noted that gradient boosting machine (GBM) have outperformed adaboost classfiers. In an empirical comparison study of supervised learning algorithms~\cite{Caruana:2006} comparing random forests and boosted decision trees, calibrated boosted trees had the best overall performance with random forests being close second. Motivated by these successes, we have chosen to use Gradient boosting tree as our classifier model. We build GBM classifiers for all the feature sets explained in Section \ref{tran} and \ref{trad} using GBM implementation inspired by~\cite{Becker:2013}, with parameters: shrinkage factor and sampling factor set to 0.5, maximum tree depth = 2 and number of iterations = 200.
\subsection{Hybrid approach}
\label{hybrid}
Investigation of the failure modes of baseline method (Haar + GBM) and CaffeNet\_FA revealed that they had failed on different images (Section \ref{sec:results}). To exploit the complementary advantages, we propose a simple scheme of averaging the spatial likelihood maps from GBMs of these two approaches and employing it in~\eqref{eqn:mlmodel}, which yields dramatic improvement.

\section{Results}
\label{sec:results}

To quantitatively evaluate the performance on 45 validation images, we used two metrics: (1) Number of localization failures - the number of images for which the dice similarity coefficient between detected kidney ROI and ground truth annotation was $<$ 0.80. (2) Detection accuracy - average dice overlap across 45 images between detection results and ground truth, which . From Table~\ref{ResultsTable}, we see that CaffeNet features without any adaptation outperformed baseline by 2\% in average detection accuracy with same number of failures. This improvement is consistent with other results reported in literature~\cite{Shie:2015}, where CaffeNet features outperform state-of-the-art pipeline. However, by allowing these network weights to get adapted to the kidney data, we achieved a performance boost of 4\% over the baseline method, with number of failure cases reducing to 10 from 12. Interestingly, tuning with the first two convolutional layers frozen yielded intermediate performance, suggesting that multiple levels of feature adaptation are important to the problem.

\begin{figure}[!ht]

	\centering
	\subfloat[Haar\label{subfig:a}]{%
		\centering
		\includegraphics[scale=0.128]{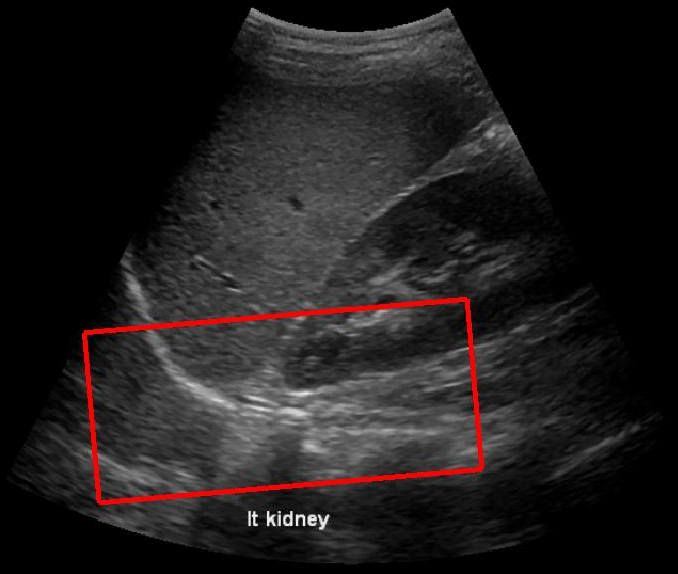}
	}
	\subfloat[CaffeNet\_FA\label{subfig:b}]{%
		\includegraphics[scale=0.143]{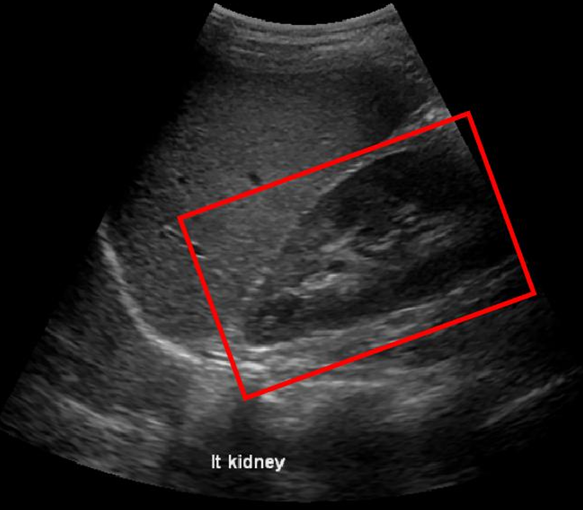}
	}          
	
	\subfloat[Haar \label{subfig:c}]{%
		\includegraphics[scale=0.13]{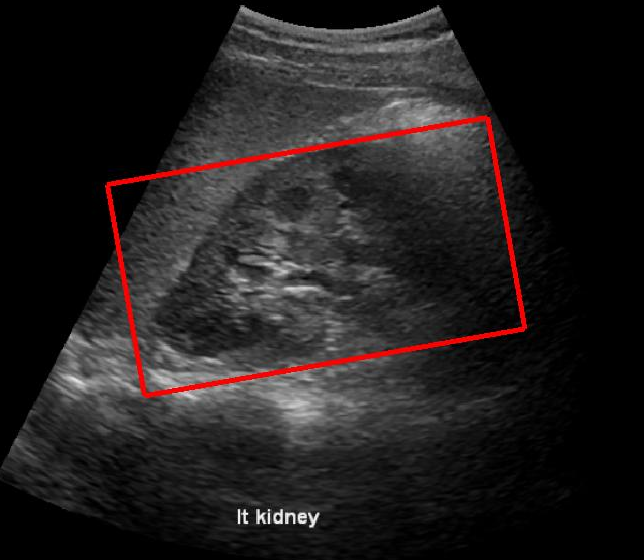}
	}
	\subfloat[CaffeNet\_FA \label{subfig:d}]{%
		\includegraphics[scale=0.128]{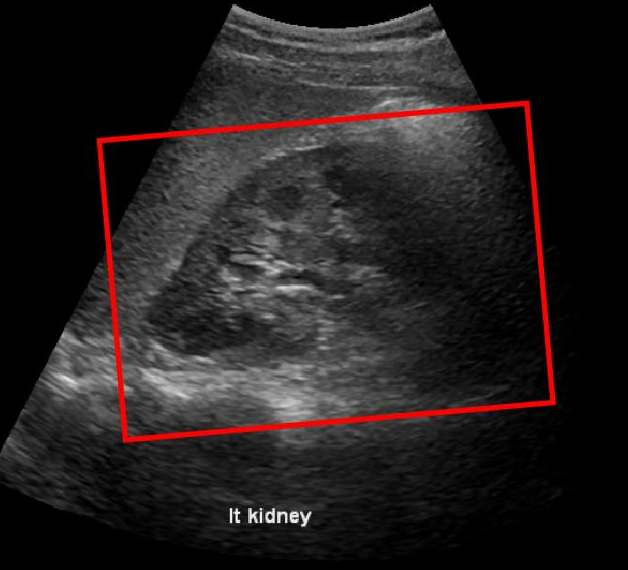}
	}
	\caption{Visual comparison of Baseline Method with CaffeNet Transfer }
	\label{fig:comparison}

\end{figure}

Fig.~\ref{fig:comparison}\subref{subfig:a} and~\subref{subfig:b} shows a case in which the baseline method was affected by the presence of diaphragm, kidney and liver boundaries creating a texture similar to renal-sinus portion, while CaffeNet had excellent localization. Fig.~\ref{fig:comparison}\subref{subfig:c} and~\subref{subfig:d} illustrate a case where CaffeNet resulted in over-segmentation containing the diaphragm, clearly illustrating that in limited data problems careful feature-engineering incorporating domain knowledge still carries a lot of relevance. Finally,  we achieved a best performance of 86\% average detection accuracy using the hybrid approach (Section \ref{hybrid}). More importantly, the number of failures of the hybrid approach was 3/45, which is 20\% better than either of the methods.  
\begin{table}[t]
\centering
	\noindent 

	\begin{tabular}{|>{\centering\arraybackslash}m{1.8cm}|>{\centering\arraybackslash}m{1.5cm}|>{\centering\arraybackslash}m{2cm}|>{\centering\arraybackslash}m{2cm}|>{\centering\arraybackslash}m{2cm}|>{\centering\arraybackslash}m{2.1cm}|}
		\hline
		\textbf{Method}&Haar Features&
	CaffeNet\_NA &
		CaffeNet\_PA &
		CaffeNet\_FA &
		{\bf Haar + CaffeNet\_FA}\\
		\hline
		\textbf{Average Dice overlap}&0.793& 0.825 & 0.831 & 0.842 & \textbf{0.857}\\
		\hline
		\textbf{\# of failures}& 12/45& 12/45& 11/45 & 10/45 & \textbf{3/45}\\
		\hline
	\end{tabular}

	\caption[performance\_comparison]{Performance Comparison on unseen 45 validation images }
			\label{ResultsTable}

\end{table}

\section{Discussion}
\label{discuss}

\begin{figure}[t]
	\centering
	\subfloat[Sample Kidney patch \label{subfig:kidney_sample}]{%
		\includegraphics[width = 0.27\textwidth, height = 0.17\textwidth]{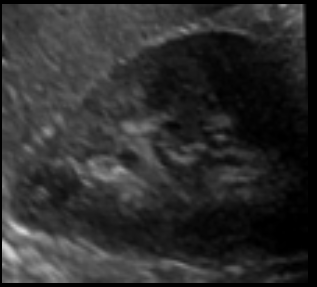}
	}
	\subfloat[Phase Congruency \label{subfig:PC}]{%
		\includegraphics[width = 0.27\textwidth, height = 0.17\textwidth]{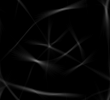}
	}
	\subfloat[Frangi Vesselness\label{subfig:frangi}]{%
		\includegraphics[width = 0.27\textwidth, height = 0.17\textwidth]{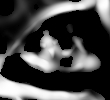}
	}\\ 
	\subfloat[CaffeNet\_NA L1\_1\label{subfig:CNNA_L11}]{%
		\includegraphics[width = 0.27\textwidth, height = 0.17\textwidth]{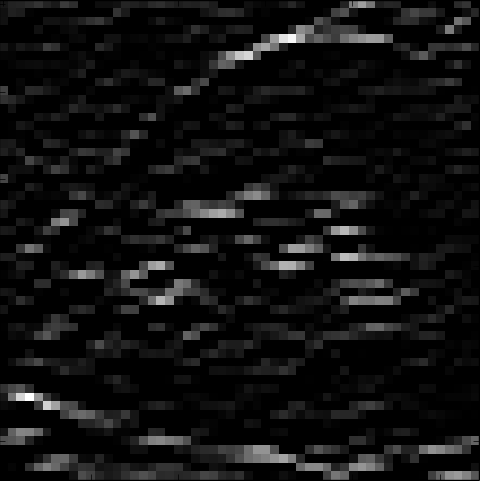}
	}
	\subfloat[CaffeNet\_NA L1\_2\label{subfig:CNNA_L12}]{%
		\includegraphics[width = 0.27\textwidth, height = 0.17\textwidth]{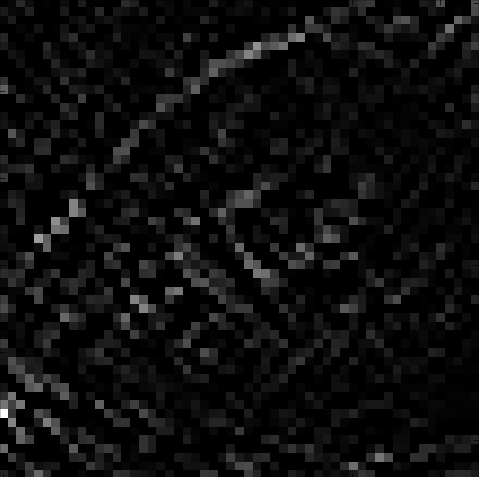}
	}
	\subfloat[CaffeNet\_NA L2\_1\label{subfig:CNNA_L2}]{%
		\includegraphics[width = 0.27\textwidth, height = 0.17\textwidth]{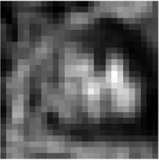}
	}\\      
    
	\subfloat[CaffeNet\_FA L1\_1\label{subfig:CNFA_L11}]{%
		\includegraphics[width = 0.27\textwidth, height = 0.17\textwidth]{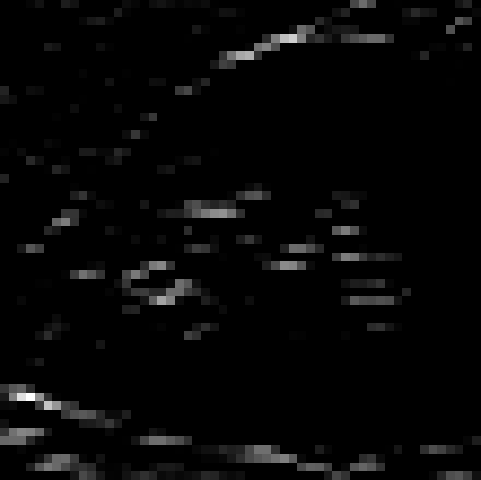}
	}
	\subfloat[CaffeNet\_FA L1\_2\label{subfig:CNFA_L12}]{%
		\includegraphics[width = 0.27\textwidth, height = 0.17\textwidth]{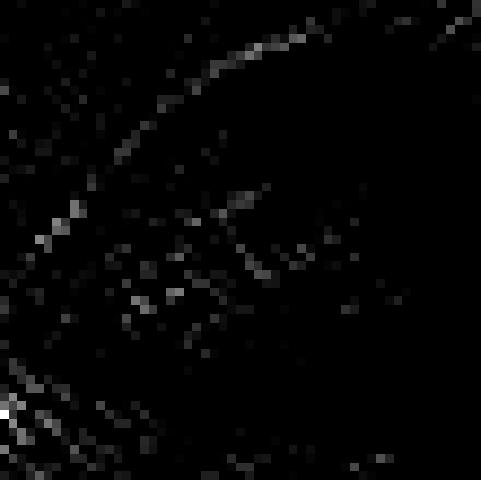}
	}
	\subfloat[CaffeNet\_FA L2\_1\label{subfig:CNFA_L2}]{%
		\includegraphics[width = 0.27\textwidth, height = 0.17\textwidth]{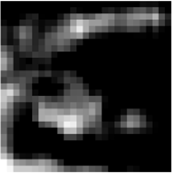}
	}\\          
	
	\caption{Filter responses for learned \& hand-engineered features for a sample patch
	        \qquad (CaffeNet\_FA(NA) LX\_Y denotes response image from  $Y^{th}$ filter of layer $X$)
	}
	\label{fig:CNVsIP}

\end{figure}

It is indeed very interesting to see that features learnt on camera images were able to outperform careful feature engineering on sharply different detection problems, in modalities whose acquisition physics are distinctly different.
Fig.~\ref{fig:CNVsIP} compares some of the response images generated from layers 1 and 2 of the learned network with traditional image processing outputs like Phase Congruency~\cite{Kovesi:2003} and Frangi vesselness filter~\cite{Frangi:1998} for an example patch. 

Here, we would like to highlight two main points: (1) Visually, we find the output has intriguing similarities with the outputs of hand crafted feature extractors optimized for Ultrasound. The response maps of Fig.~\ref{fig:CNVsIP}\subref{subfig:CNFA_L11} and \subref{subfig:CNFA_L2} are similar to~\ref{fig:CNVsIP}\subref{subfig:PC} and \subref{subfig:frangi}. This is very encouraging because of the fact that CNNs learns features that are equivalent to some of these widely used non-linear feature extractors. (2) The second important observation here is the reduction in speckle noise on CaffeNet\_FAL1\_1, compared to CaffeNet\_PAL1\_1.
By carefully tuning CaffetNet features on ultrasound data, the model was able to learn the underlying noise characteristics, while preserving edges, and this resulted in a much improved response map as shown in Fig.~\ref{fig:CNVsIP}\subref{subfig:CNFA_L11} and \subref{subfig:CNFA_L2}.
 
\begin{table}[h]
	\centering
		\begin{tabular}{|>{\centering\arraybackslash}m{5cm}|>{\centering\arraybackslash}m{1.3cm}>{\centering\arraybackslash}m{1.3cm}>{\centering\arraybackslash}m{1.3cm}>{\centering\arraybackslash}m{1.3cm}>{\centering\arraybackslash}m{1.3cm}|}
		
		\hline
		
		&&&{\bf Layer}&&\\
			\cline{2-6}
		\centering  & \centering Conv1 & \centering Conv2 & \centering Conv3 & \centering Conv4 & \centering Conv5 \tabularnewline
		\hline
		\centering  {\bf \# of filters with $\geq$ 40\% change} & \centering 0 & \centering 5 & \centering 125 & \centering 22 & \centering 62 \tabularnewline
		\hline
	\end{tabular}
	\caption{No. of filters in each layer that changed by more than 40\% in $\ell_2$ norm}
		\label{filterChangeTable}	
\end{table}

Further, we quantitatively analyzed changes (\% in $\ell_2$ norm) in filter weights in each layer to identify significant trends. Table~\ref{filterChangeTable} shows a large number of filters have significantly changed in the 3rd layer, with filters in the 1st and 2nd layer showing minimal change. This is possibly due to the lower level features being fairly the same for both natural and ultrasound images. We also noted that the use of ReLU as the activation function also avoided the vanishing gradient problem, resulting in this skew in distribution of weight changes across layers. The response images past layer 2 proved to be difficult to interpret, and may require more intensive techniques.   Our quantitative results and the literature in the field show that a great deal of the power of deep networks lies in these layers, and so we feel this is an important area for our future investigation.

In a clinical context, the interpretability of models is crucial and we feel this insight into why the deep CNN was able to outperform hand-crafted features is as important as the results demonstrated in Sec.~\ref{sec:results}. We also see this as opening up new ways of understanding and utilizing deep networks for medical problems.

\bibliographystyle{plain}
\bibliography{bibliographyv2}

\end{document}